\pdfoutput=1
\documentclass[11pt]{article}

\usepackage[final]{acl}
\usepackage{booktabs}
\usepackage{times}
\usepackage{latexsym}

\usepackage[T1]{fontenc}

\usepackage[utf8]{inputenc}

\usepackage{microtype}

\usepackage{inconsolata}

\usepackage{graphicx}

\usepackage{hyperref}
\usepackage{subcaption}
\usepackage{adjustbox}
\usepackage{graphicx}
\usepackage{amssymb}
\usepackage{algorithm}
\usepackage{algpseudocode}
\usepackage{booktabs}
\usepackage{amsthm, amscd, amsfonts, multirow, array}
\usepackage{makecell}
\usepackage{IEEEtrantools}
\usepackage{amsfonts}
\usepackage{amsmath}
\usepackage{enumitem}
\usepackage{colortbl}
\usepackage{textcomp}
\usepackage{makecell}
\usepackage{tabularx}
\usepackage[capitalize,noabbrev]{cleveref}

\newcommand{\email}[1]{\href{mailto:#1}{\texttt{#1}}}

\newlength\savewidth

\title{TASO: Task-Aligned Sparse Optimization for Parameter-Efficient Model Adaptation}

\author{
Daiye Miao\textsuperscript{1}\thanks{~~Equal contribution.},
Yufang Liu\textsuperscript{1}\footnotemark[1],
Jie Wang\textsuperscript{1},
Changzhi Sun\textsuperscript{1},\\
\textbf{Yunke Zhang}\textsuperscript{2},
\textbf{Demei Yan}\textsuperscript{2},
\textbf{Shaokang Dong}\textsuperscript{2},
\textbf{Qi Zhang}\textsuperscript{3},
\textbf{Yuanbin Wu}\textsuperscript{1} \\
\textsuperscript{1}East China Normal University \quad
\textsuperscript{2}Honor Device Co., Ltd. \quad
\textsuperscript{3}Fudan University \\
\email{dymiao@stu.ecnu.edu.cn} \quad 
\email{yfliu.antnlp@gmail.com} \quad
\email{ybwu@cs.ecnu.edu.cn}
}

\begin{document}
\maketitle
\begin{abstract}
LoRA has become one of the most widely used parameter-efficient fine-tuning methods due to its simplicity and effectiveness. However, numerous studies have shown that LoRA often introduces substantial parameter redundancy, which not only increases the number of trainable parameters but also hinders the effectiveness of fine-tuning. Since identifying redundant parameters in LoRA is inherently difficult, how to eliminate them efficiently and accurately remains a challenging problem. 
In this paper, we propose TASO, a redundancy reduction method that leverages importance information from the pretrained model's weights to mitigate LoRA redundancy. Specifically, we estimate parameter importance on downstream tasks and identify task-specific core regions based on the distribution of importance scores. The location information of these core regions is then used to determine the sparse structure of LoRA modules, enabling redundancy removal before fine-tuning. 
Our approach significantly reduces the number of trainable parameters required for task adaptation, while providing a novel task-aligned perspective for LoRA redundancy reduction. Experimental results demonstrate that, with a parameter budget comparable to LoRA with rank $r = 1$, TASO consistently outperforms standard LoRA across multiple tasks, achieving strong fine-tuning performance while effectively eliminating redundant parameters.
Code is available at  \href{https://github.com/123bigmirros/TASO.git}{https://github.com/123bigmirros/TASO.git}.
\end{abstract}

\section{Introduction}

With the rapid development of large-scale pretrained language models~\cite{zhao2023survey,zhou2024comprehensive,minaee2025largelanguagemodelssurvey,zhang2024comprehensive}, efficiently adapting them to downstream tasks has become a central challenge\cite{ding2023parameter,han2024parameter,wang2024parameter}. Although full fine-tuning remains highly effective, it requires storing a separate set of model weights for each task, leading to substantial storage and computational overhead. To address this, a variety of Parameter-Efficient Fine-Tuning (PEFT) methods\cite{zhou-etal-2024-empirical,prottasha2025peft} have been proposed, which aim to adapt models by updating only a small subset of parameters. Among them, Low-Rank Adaptation (LoRA) has gained significant popularity due to its simplicity and efficiency~\cite{hu2022lora}. LoRA freezes the original model weights and injects trainable low-rank matrices, thereby avoiding additional inference overhead and becoming one of the most widely adopted PEFT techniques in practice.

However, recent studies have revealed substantial redundancy not only in the delta parameters generated during fine-tuning, but also within the LoRA modules themselves~\cite{yu2024languagemodelssupermario,panda2024lotteryticketadaptationmitigating,kopiczko2024veravectorbasedrandommatrix,zhang2023adaloraadaptivebudgetallocation}, motivating further efforts to reduce the number of trainable parameters without sacrificing performance. Yet, most existing approaches still require training a complete LoRA module before gradually pruning unimportant weights during training, which fails to fundamentally reduce training costs.

Inspired 
by recent findings on the spatial concentration of task-relevant core regions in pretrained models~\cite{zhang-etal-2024-unveiling-linguistic}, we propose a novel method to sparsify LoRA prior to fine-tuning. 
Specifically, we estimate the importance of pretrained weights by leveraging downstream task gradients to design a sparsified structure for the low-rank adaptation matrices. Building on the observed clustering patterns of high-importance weights, we pre-define zero positions for low-importance regions, thereby constructing sparse LoRA modules. During training, these modules dynamically aggregate updates only to the non-zero positions, ensuring efficient adaptation while preserving the pre-defined sparse structure.

Unlike methods that perform pruning during training, our approach eliminates redundancy in advance, based on the inherent sparsity of delta weights and the localized importance of core parameters. By identifying the subset of parameters to be trained before fine-tuning begins, we ensure that updates are concentrated on the most critical regions of the original model, thereby significantly reducing parameter overhead while maintaining model effectiveness.

Experimental results show that our method consistently outperforms standard high-rank LoRA on multiple benchmarks, even when using a parameter budget comparable to rank-1 LoRA. These findings confirm the effectiveness of proactively identifying and leveraging key parameter regions, enabling substantial reductions in fine-tuning cost without compromising generalization. 
The summary of our contributions is as follows:
\begin{itemize}[topsep=4pt, noitemsep,leftmargin=*]
    \item We introduce an importance-guided LoRA sparsification method that significantly reduces the number of trainable parameters without increasing inference cost, offering a new and practical solution for efficient adaptation of large language models.
    \item We introduce a sparsity-aware learning rate scaling strategy that ensures the retained parameters in pruned LoRA modules can be fully optimized, thereby preserving downstream adaptation performance.
    \item TASO consistently outperforms LoRA with significantly fewer trainable parameters across various NLP tasks and model architectures.
\end{itemize}
\section{Related Work}

\subsection{Parameter-Efficient Fine-Tuning Methods}
Parameter-efficient fine-tuning (PEFT) aims to adapt large pre-trained models by optimizing only a small subset of parameters while keeping the majority of the model frozen. Representative methods include Adapter tuning~\cite{houlsby2019parameterefficienttransferlearningnlp}, Prefix and Prompt Tuning~\cite{li2021prefixtuningoptimizingcontinuousprompts,li-liang-2021-prefix},BitFit~\cite{ben-zaken-etal-2022-bitfit},IA3~\cite{liu2022few} which introduce lightweight modules or continuous prompts into the model to enable efficient task adaptation. Another direction focuses on identifying and updating specific subsets of the model’s parameters~\cite{hu2022sparsestructuresearchparameterefficient,ben-zaken-etal-2022-bitfit,guo-etal-2021-parameter}.

LoRA remains one of the most widely used fine-tuning methods due to its simplicity and effectiveness. Building on its success, many studies propose extensions to improve its flexibility or generalization. Some works explore combining LoRA modules across tasks or styles~\cite{luo2024moeloracontrastivelearningguided,huang2024lorahubefficientcrosstaskgeneralization,shah2023ziplorasubjectstyleeffectively,hu2022sparsestructuresearchparameterefficient}. Others focus on modifying its internal structure~\cite{tian2024hydraloraasymmetricloraarchitecture,DBLP:conf/iclr/KopiczkoBA24,DBLP:journals/corr/abs-2405-17604}. These works demonstrate that while LoRA is highly efficient, further improvements can be achieved through better module design and composition strategies.

\subsection{Redundancy in LoRA Parameters}

Although LoRA significantly reduces the number of trainable parameters, recent studies have shown that it can still introduce redundancy.  
For example, \citet{yu2024languagemodelssupermario} show that LoRA’s delta parameters still contain substantial redundancy, and demonstrate that techniques originally developed for other modules remain effective when applied to LoRA.

To address this, methods like AdaLoRA~\cite{zhang2023adaloraadaptivebudgetallocation} and DyLoRA~\cite{valipour2023dyloraparameterefficienttuning} dynamically allocate rank across layers by pruning unimportant directions.  
Other approaches impose structural sparsity on LoRA matrices.  
SoRA~\cite{ding2023sparselowrankadaptationpretrained} introduces trainable gates to zero out redundant components during training, while RoseLoRA~\cite{wang2024roselorarowcolumnwisesparse} enforces row- and column-wise sparsity for more targeted adaptation.

These works demonstrate that reducing LoRA redundancy—either by pruning or structured sparsity—can maintain or even improve performance while further lowering the adaptation cost.  
Our method continues along this line, proposing a more effective strategy for eliminating unnecessary parameters in LoRA modules.



\section{Background}
\begin{figure*}[t]
    \centering
    \includegraphics[width=0.9\textwidth]{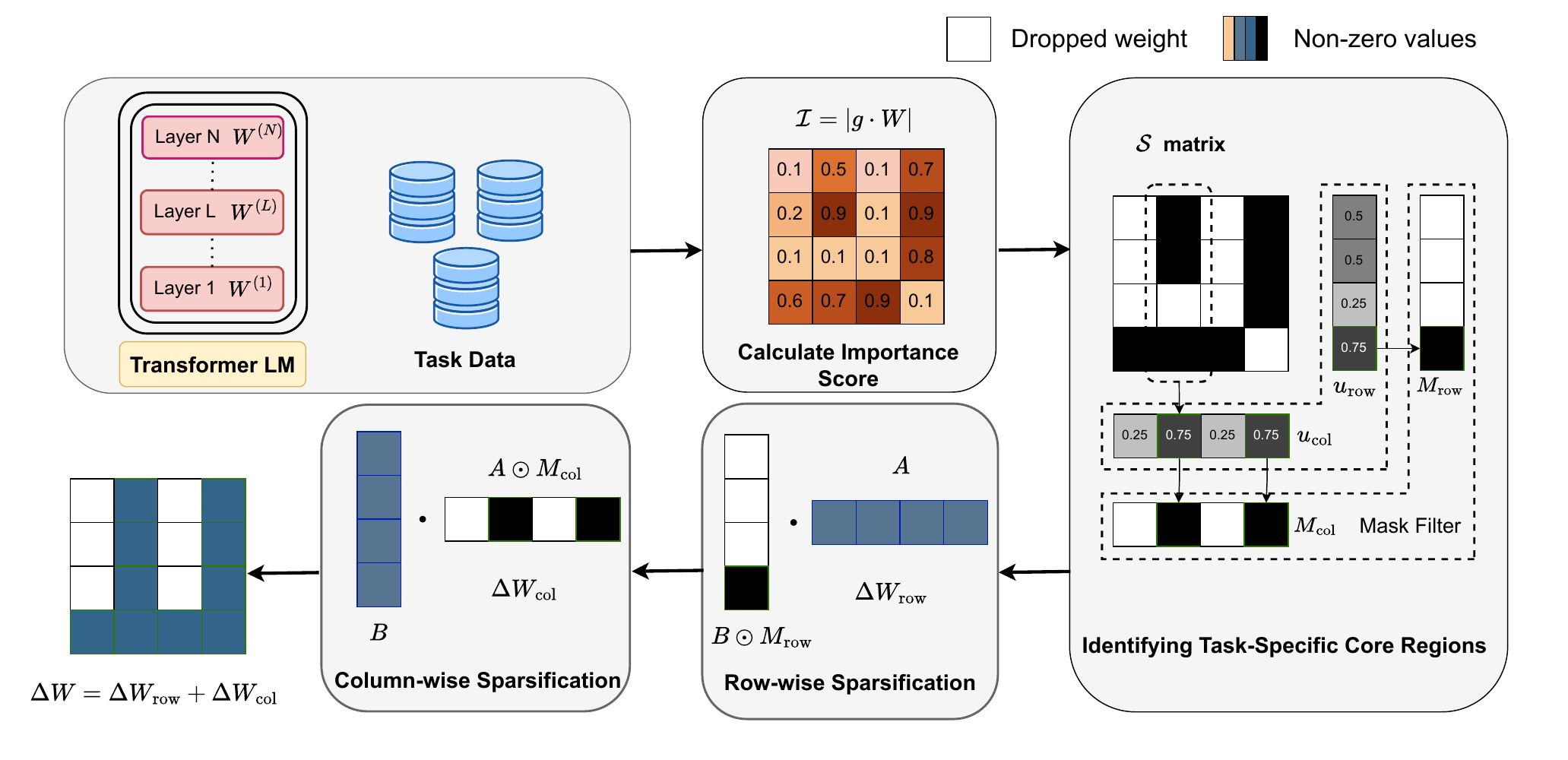}
    \caption{
    Overview of TASO. We compute task-specific importance from SFT loss to determine core regions, perform structured pruning of LoRA, and scale learning rate based on sparsity level.
    }
    \label{fig:method_overview}
\end{figure*}
\paragraph{Low-rank Adaptation}
Before introducing our approach,
we first briefly recap low-rank adaptation (LoRA), which operates by freezing the pre-trained weight matrices $\mathbf{W}_0 \in \mathbb{R}^{p \times q}$ while introducing trainable low-rank decomposition matrices to model weight updates. Specifically, the weight update $\Delta\mathbf{W}$ is factorized into two low-rank matrices $\mathbf{A} \in \mathbb{R}^{r \times q}$ and $\mathbf{B} \in \mathbb{R}^{p \times r}$ such that $\Delta\mathbf{W} = \mathbf{BA} \in \mathbb{R}^{p \times q}$, where $r \ll \min(p,q)$ represents the intrinsic rank dimension controlling the adaptation capacity. 

The modified forward pass of a layer can be expressed as:
\begin{equation}
    \mathbf{y} \leftarrow \mathbf{W}_0\mathbf{x} + \mathbf{BAx},
\end{equation}
where $\mathbf{x}$ is the input and $\mathbf{y}$ is the output of the layer. The key computational advantage stems from the low-rank projection $\mathbf{z} \leftarrow \mathbf{BAx}$, which enables efficient adaptation while maintaining the original model structure. The hyperparameter $r$ governs both the number of trainable parameters and the expressiveness of the adaptation. Existing approaches~\cite{ding2023sparselowrankadaptationpretrained,zhang2023adaloraadaptivebudgetallocation} often train full-rank LoRA parameters and subsequently prune or reduce them, which may incur unnecessary training overhead. In contrast, we aim to explore a more efficient alternative that reduces adaptation cost from the outset.

\paragraph{Parameter Importance Estimation}
Quantifying the relative importance of model parameters constitutes a fundamental challenge in neural network analysis, with broad implications for understanding and optimizing model behavior~\cite{DBLP:conf/acl/DaiDHSCW22,DBLP:conf/emnlp/YaoGLZWWZ24}. By identifying parameters that are critical for maintaining task performance, this analysis enables more principled approaches to model adaptation and refinement~\cite{wang2024roselorarowcolumnwisesparse}. A theoretically grounded framework for parameter importance estimation derives from sensitivity analysis, where importance is defined as the element-wise product of a parameter's value and its corresponding loss gradient. Formally, for each parameter $\theta_i$, the importance metric is given by:
\[
\mathcal{I}_i(\theta) = \left| \theta_i \cdot \frac{\partial \mathcal{L}}{\partial \theta_i} \right|.
\]
This formulation simultaneously captures a parameter's magnitude and its influence on the loss landscape. High-scoring parameters are essential to model performance, while low-scoring ones can be modified with minimal impact. The method provides an efficient way to evaluate parameter salience across learning scenarios.

\section{Method}
The proposed methodology is illustrated in Figure~\ref{fig:method_overview}. Our approach consists of three key stages: First, we identify task-specific core regions within the model by estimating the importance of individual weights with respect to the target task. Subsequently, through analysis of the distribution patterns among highly important weights, we observe distinct row- and column-wise clustering characteristics. This empirical observation informs the design of an optimized sparsified LoRA structure. Finally, we perform fine-tuning based on this task-aligned sparse architecture to enhance parameter efficiency while maintaining model performance.

\subsection{Identifying Task-Specific Core Regions}



\begin{figure*}[t]
    \centering
    \includegraphics[width=0.9\textwidth]{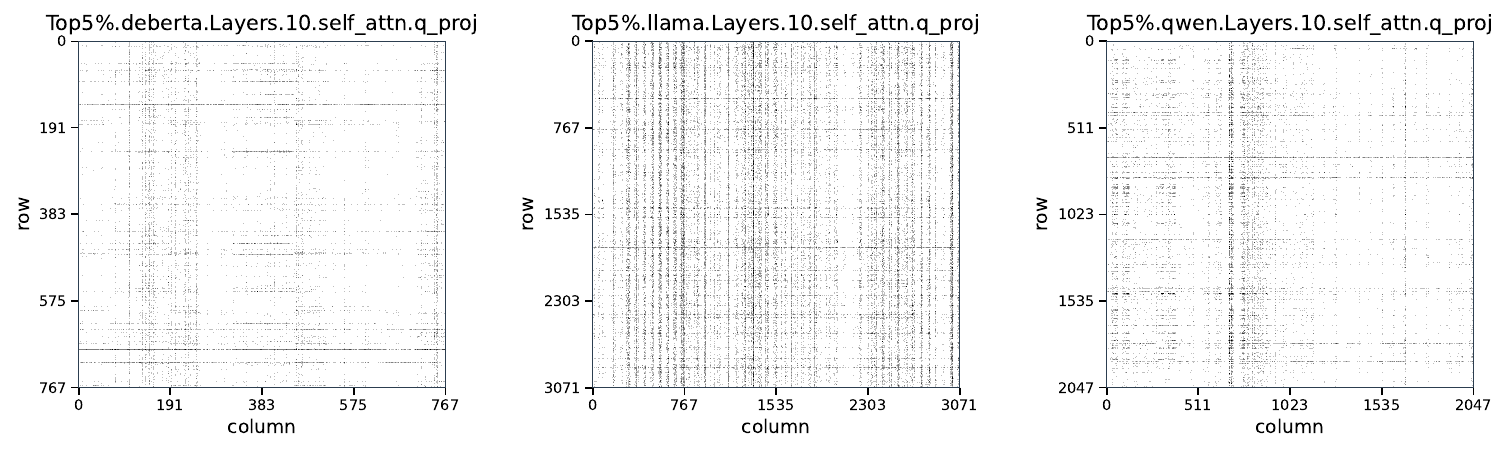}
    \caption{
    Visualization of task-specific important weights identified by sensitivity analysis.  
    Heatmaps show the top-5\% important parameters in the self-attention query matrices of  
    DeBERTa-v3 on RTE, LLaMA3.2 3B on GSM8K, and Qwen2.5 3B on GSM8K.  
    The important weights concentrate in specific rows and columns of the matrices.
    }
    \label{fig:mask_visualization}
\end{figure*}

Recent studies~\cite{zhang-etal-2024-unveiling-linguistic, DBLP:conf/aaai/HuangHJZC25} have demonstrated that parameter contributions in pre-trained models are highly task-dependent. Notably, \citet{zhang-etal-2024-unveiling-linguistic} discover that specific model capabilities are governed by sparse subsets of weights (termed ``linguistic regions''), while the majority of parameters remain inactive. This observation motivates our hypothesis that LoRA updates could achieve greater efficiency by concentrating on task-relevant parameter subspaces.
Given a training task, we compute the importance score $\mathcal{I}_i$ for each parameter $\theta_i$. We then define a binary mask $\mathcal{S}$, where:
\begin{equation}
\mathcal{S}_i = 
\begin{cases}
1, & \text{if } \mathcal{I}_i \text{ ranks in top-}k\% \\
0, & \text{otherwise}
\end{cases}
\end{equation}
Here, $\mathcal{S}_i$ indicates whether parameter $\theta_i$ is among the top-$k\%$ most important parameters for the task.

Figure~\ref{fig:mask_visualization} highlights the top-5\% most important parameters in the self-attention query matrices of three models: DeBERTa-v3 (on RTE), LLaMA3.2 3B (on GSM8K), and Qwen2.5 3B (on GSM8K). 
The analysis reveals two key findings: (1) Important parameters exhibit non-uniform, clustered distributions with clear row-wise and column-wise patterns; (2) These structural regularities persist across diverse architectures and tasks, suggesting an inherent organization of parameter importance in pre-trained models.
The observed patterns motivate our pruning strategy for LoRA parameters, where we selectively preserve these structurally important dimensions while eliminating redundant ones before training, thereby significantly reducing both computational overhead and memory requirements during adaptation.

To unify the importance of rows and columns, we compute average importance scores for each row $i$ and column $j$ from the binary mask $\mathcal{S}$:
\[ 
u_i^{\text{row}} = \frac{1}{m}\sum_{j=1}^m \mathcal{S}_{i,j}, \quad 
u_j^{\text{col}} = \frac{1}{n}\sum_{i=1}^n \mathcal{S}_{i,j}
\]
which represent the density of important parameters in each dimension.
We then perform a global ranking  based on importance scores $u$,  
and select the top-$p\%$ entries as the task-specific core region.
Among them, we define $\mathcal{R}_{\text{core}}$ as the set of row indices and  $\mathcal{C}_{\text{core}}$ as the set of column indices.
These core regions will guide structured pruning in LoRA modules.

\subsection{Structured LoRA Pruning Guided by Core Regions}

Building upon the identified core regions $\mathcal{R}_{\text{core}}$ and $\mathcal{C}_{\text{core}}$, we implement a structured pruning strategy for LoRA parameters that optimizes training efficiency. To minimize training overhead, we set the rank $r=1$. The approach combines rank-constrained decomposition with mask-based sparsification, which includes two sequential stages: row sparsification followed by column sparsification.

\paragraph{Row-wise Sparsification} First, we apply row-wise sparsification to the LoRA matrix $A$. This is implemented by masking all rows not contained in $\mathcal{R}_{\text{core}}$, effectively preserving only the parameters from the identified important rows. The training objective during this phase can be expressed as:
\[
\Delta \mathbf{W}_{\text{row}} = \mathbf{B} \cdot (\mathbf{A} \odot \mathbf{M}_{\text{row}}),
\]
where $\mathbf{M}_{\text{row}}$ is the row mask matrix defined by $\mathbf{M}_{\text{row}}(i,:) = \mathbb{I}(i \in \mathcal{R}_{\text{core}})$ for each row $i$, and $\odot$ denotes element-wise multiplication.
Row-wise sparsification restricts updates to rows in $\mathcal{R}_{\text{core}}$ while preserving global column propagation.
Following row-wise training, we update the original parameters by accumulating the learned delta weights:
\[
\mathbf{W}_{\text{updated}} = \mathbf{W}_0 + \Delta \mathbf{W}_{\text{row}}.
\]

\paragraph{Column-wise Sparsification} Subsequently, we perform column-wise sparsification on matrix $B$ using the same approach. The column mask $M_{\text{col}}$ preserves only columns from $\mathcal{C}_{\text{core}}$, with the training operation formulated as:
\[
\Delta \mathbf{W}_{\text{col}} = (\mathbf{B} \odot \mathbf{M}_{\text{col}}) \cdot \mathbf{A}.
\]
Similarly, column-wise sparsification restricts updates to columns in $\mathcal{C}_{\text{core}}$ while preserving global row projection.

The complete parameter update then becomes:
\[
\mathbf{W}_{\text{final}} = \mathbf{W}_{\text{updated}} + \Delta \mathbf{W}_{\text{col}}.
\]
This formulation ensures that $\Delta \mathbf{W}$ introduces non-zero updates exclusively within the task-specific core region,  
achieving targeted sparsification while maintaining the expressiveness of the rank-1 LoRA adaptation.

\subsection{Scaling Learning Rates under Structured Sparsity}



Structured pruning significantly reduces the number of trainable parameters in LoRA modules, thereby impairing the model's adaptation capability during fine-tuning. To compensate for this sparsity, the DARE method~\cite{yu2024languagemodelssupermario} proposes parameter rescaling:
\begin{equation}
\Delta \mathbf{W}_{\text{rescaled}} = \frac{1}{1-\rho} \cdot \Delta \mathbf{W}_{\text{pruned}}
\end{equation}
where $\Delta \mathbf{W}_{\text{pruned}}$ denotes the parameter matrix after structured pruning (with some elements zeroed out), and $\rho \in [0,1)$ represents the pruning ratio (fraction of zeroed elements~\footnote{The value of $\rho$ is determined by calculating the proportion of non-zero elements in the $\Delta W$.}). The remaining non-zero parameters are amplified by a factor of $1/(1-\rho)$ to maintain the original magnitude scale.



Unlike conventional post-hoc rescaling of $\Delta \mathbf{W}$, our approach mitigates the effects of sparsity by adjusting the learning rate. This avoids potential numerical instability caused by parameter scaling.
Specifically, we scale the LoRA factors as:
\begin{equation}
\Delta \mathbf{W} = \left( \sqrt{\tfrac{1}{1-\rho}} \cdot \mathbf{A} \right) \left( \sqrt{\tfrac{1}{1-\rho}} \cdot \mathbf{B} \right) = \tfrac{1}{1-\rho} \cdot \mathbf{A} \mathbf{B},
\end{equation}
which corresponds to scaling the learning rate as:
\begin{equation}
\text{lr}_{\text{scaled}} = \sqrt{\tfrac{1}{1-\rho}} \cdot \text{lr},
\end{equation}
here the scaling factor compensates for gradient attenuation due to sparsity.

This approach preserves the desired effect of sparsity compensation,  
while providing finer control over LoRA's impact on the model's overall behavior.  
It also avoids abrupt changes to $\Delta \mathbf{W}$, leading to more stable fine-tuning.

\definecolor{mygreen}{RGB}{75,139,59} 
\section{Results}


\paragraph{Datasets and Models}

We conduct comprehensive evaluations across two major categories of language models. For \textbf{decoder-only} language models, we evaluate multiple benchmarks: mathematical reasoning (GSM8K~\cite{cobbe2021training}), question answering (BoolQ~\cite{clark2019boolq}), word sense disambiguation (WiC~\cite{pilehvar2018wic}), and scientific question answering (ARC~\cite{clark2018think}, including ARC-e and ARC-c). We use Qwen2.5-3B~\cite{qwen2025qwen25technicalreport} and LLaMA3.2-3B~\cite{grattafiori2024llama3herdmodels} as backbone models, splitting each task's validation set into development and test sets (2:8 ratio).
For \textbf{encoder-based} models, we evaluate on GLUE~\cite{wang2018glue}, covering tasks like grammatical acceptability (CoLA~\cite{warstadt2019neural}), sentiment analysis (SST-2~\cite{socher2013recursive}), paraphrase identification (MRPC~\cite{dolan2005automatically}, QQP~\cite{quora2017question}), sentence similarity (STS-B~\cite{cer2017semeval}), and natural language inference (MNLI~\cite{williams2018broad}, QNLI~\cite{rajpurkar2016squad}, RTE~\cite{dagan2005pascal}). DeBERTaV3-base~\cite{he2021debertav3} serves as the backbone model.

\begin{table*}[t]
\centering
\resizebox{0.95\textwidth}{!}{
\begin{tabular}{lccc c c c c c}
\toprule
\textbf{Method} & \textbf{Model} & \textbf{ \#Param.} & \textbf{GSM8K} & \textbf{BoolQ} & \textbf{WiC} & \textbf{ARC-e} & \textbf{ARC-c} & \textbf{Avg.} \\
\midrule
Fine-tune       & \multirow{10}{*}{Qwen2.5 3B} & 3151.91\,M & 67.20 & 87.76 & 72.35 & 93.74 & 82.72 & 80.75 \\
IA3             &                              &   1.35\,M & 64.64 & 88.26 & 72.54 & 93.16 & 81.76 & 80.07 \\
Adapter         &                              &  67.10\,M & 66.44 & 88.22 & 71.56 & 93.84 & 81.98 & 80.41 \\
LoRA (r=8)      &                              &  14.97\,M & 68.83 & 87.50 & 73.13 & 94.58 & 82.08 & 81.22 \\
LoRA (r=32)     &                              &  59.87\,M & 68.34 & 87.95 & 72.94 & 94.63 & 82.83 & 81.34 \\
DoRA (r=32)     &                              &  65.98\,M & 68.53 & 88.37 & \textbf{74.11} & 94.73 & 83.26 & 81.80 \\
AdaLoRA (r=32)  &                              &  61.32\,M & 68.72 & \textbf{89.14} & 73.31 & 94.63 & 81.66 & 81.09 \\
VERA (r=1024)   &                              &  1.42\,M & 68.90 & 85.43 & 71.56 & 94.05 & 81.02 & 80.19 \\
LoRA-XS (r=128) &                              &  4.13\,M & 69.09 & 82.07 & 73.52 & 92.89 & 84.00 & 80.31 \\
TASO            &                              &   2.06\,M & \textbf{70.04} & 88.64 & 73.33 & \textbf{94.84} & \textbf{85.50} & \textbf{82.47} \\
\hline
Fine-tune       & \multirow{10}{*}{LLaMA3.2 3B} & 3266.58\,M & \textbf{45.11} & 80.73 & 74.21 & 85.11 & 70.46 & 71.12 \\
IA3             &                              &   0.91\,M & 32.89 & 84.51 & 70.98 & 84.79 & 67.59 & 68.15 \\
Adapter         &                              &  50.32\,M & 36.39 & 85.28 & 72.54 & 86.53 & 66.63 & 69.47 \\
LoRA (r=8)      &                              &  12.16\,M & 35.45 & 80.73 & 71.76 & 85.21 & 64.60 & 67.55 \\
LoRA (r=32)     &                              &  48.63\,M & 39.60 & 82.14 & 71.96 & 86.53 & 67.48 & 69.54 \\
DoRA (r=32)     &                              &  53.73\,M & 40.00 & 84.63 & 70.58 & 86.32 & 68.44 & 70.00 \\
AdaLoRA (r=32)  &                              &  49.51\,M & 41.06 & 87.80 & 73.92 & 85.58 & 68.76 & 71.02 \\
VERA (r=1024)   &                              &  1.09\,M & 38.00 & 85.85 & \textbf{74.50} & 84.27 & 66.73 & 69.87 \\
LoRA-XS (r=128) &                              &  3.21\,M & 33.74 & 79.89 & 72.94 & 84.06 & 67.37 & 67.20 \\
TASO            &                              &   1.67\,M & 39.81 & \textbf{88.53} & 74.50 & \textbf{86.74} & \textbf{68.86} & \textbf{71.82} \\
\bottomrule
\end{tabular}
}
\caption{
Comparison of full fine-tuning and PEFT methods on decoder-only models Qwen2.5-3B and LLaMA3.2-3B.  
\#Params. denotes the number of trainable parameters used during task-specific tuning.  
We evaluate methods across diverse task types, including reasoning, classification, and understanding.  
Avg. indicates the average accuracy across all evaluated tasks.
}
\label{tab:LM_cmp}
\end{table*}

\paragraph{Baselines}
We compare with the following baselines:
\begin{itemize}[topsep=4pt, noitemsep,leftmargin=*]
    \item \textbf{Fine-tune}: Updates all model parameters during task-specific training, serving as the conventional upper-bound baseline.
    \item \textbf{BitFit}: A highly parameter-efficient approach that optimizes only the bias terms while freezing other parameters.
    \item \textbf{Adapter}: Inserts lightweight bottleneck modules between transformer layers, preserving original model weights through complete freezing.
    \item \textbf{LoRA}: Performs low-rank decomposition of weight matrices, injecting trainable rank-defficient matrices for efficient adaptation.
    \item \textbf{IA3}: Implements learnable scaling vectors on attention activations and feed-forward network outputs for localized adaptation. 
    \item \textbf{DoRA}: Enhances LoRA by decoupling weight updates into directional and magnitude components, improving both expressiveness and training stability.
\end{itemize}
All baselines are implemented with standard configurations to ensure fair comparison.

\paragraph{Evaluation Metrics}
For decoder-based models, we report zero-shot accuracy on GSM8K, BoolQ, WiC, and ARC.For benchmarks that only provide validation sets , we split the validation set into 20\% development and 80\% testing. The development split is used for hyperparameter tuning, while the testing split is reserved exclusively for final evaluation  
For encoder-based models, we follow GLUE standard metrics: Matthews correlation coefficient for CoLA, accuracy for SST-2, MRPC, QNLI, and RTE, matched accuracy for MNLI, accuracy and F1 for QQP, and Pearson and Spearman correlations for STS-B.

\paragraph{Implementation Details}
Encoder experiments are conducted on NVIDIA RTX 3090 GPUs, while decoder experiments are run on NVIDIA A100 GPUs.  
We provide the learning rate settings for all methods, including LoRA (decoder and encoder variants), in Appendix~\ref{appendix:learning-rates}.
For task-specific importance estimation, we set \( k = 5\% \), selecting the top 5\% of parameters based on their importance scores.  
To identify the task-specific core region, we compute row- and column-wise retention ratios, jointly rank them, and set \( p = 5\% \) to extract the top 10\% entries.




\subsection{Results}
\begin{table*}[t]
\centering
\resizebox{\textwidth}{!}{
\begin{tabular}{lcc c c c c c c cc}
\toprule
\textbf{Method} & \textbf{\#Param.} & \textbf{CoLA} & \textbf{MRPC} & \textbf{QNLI} & \textbf{QQP} & \textbf{RTE} & \textbf{SST-2} & \textbf{STS-B} & \textbf{MNLI} & \textbf{Avg.} \\
\midrule
Fine-Tune         & 184 M   & 69.21 & 89.22 & 93.78 & 92.05/89.31 & 82.49 & 95.64 & 91.59 & 89.98/89.95 & 87.82 \\
Adapter           & 1.41 M  & 69.00 & 89.90 & 93.79 & 91.45/88.88 & 82.44 & 95.16 & 91.45 & 90.11/90.11 & 87.85 \\
Bitfit            & 0.1 M   & 68.70 & 87.16 & 91.90 & 87.86/84.20 & 76.12 & 94.38 & 89.71 & 87.45/87.45 & 85.18 \\
LoRA (r=8)        & 1.33 M  & 69.73 & 89.71 & 93.76 & 91.95/89.26 & 85.32 & 95.57 & 91.86 & 90.47/90.46 & 88.38 \\
LoRA (r=16)       & 2.65 M  & 69.87 & 89.91 & 93.46 & 92.22/89.63 & 87.05 & 95.53 & 91.79 & 90.55/90.31 & 88.62 \\
SoRA             &    1.33 M    &   71.48   &   \textbf{91.98}    &   \textbf{94.28}   &    \textbf{92.39}/\textbf{89.87}     & 87.77      &   95.64    &   92.22    &     90.35/90.38        &  89.36      \\
TASO              & 0.18 M  & 69.86 & 90.03 & 93.72 & 91.30/88.57 & 88.01 & 95.60 & 92.30 & 90.36/90.44 & 88.73 \\


TASO$^\dag$             &     0.75 M        &    \textbf{71.69}    &   91.83   &   94.17    &   91.60/88.94   &      \textbf{88.49}       &     \textbf{95.87}  &    \textbf{92.40}   &   \textbf{90.61}/\textbf{90.89}    &    \textbf{89.43}     \\

\bottomrule
\end{tabular}
}
\caption{
Comparison of full fine-tuning and PEFT methods on the encoder-only model DeBERTa-v3-base.  
Total Params indicates the number of trainable parameters during fine-tuning.  
We evaluate performance across multiple GLUE tasks of various types.  
Avg. denotes the average score over all tasks.
TASO$^\dag$ performs 4 rounds of iterative fine-tuning, where the incrementally trained weights from each round are merged into the model.
}
\label{tab:BERT_cmp}
\end{table*}
The experimental results of parameter-efficient fine-tuning methods across different model architectures are systematically presented in Tables~\ref{tab:LM_cmp} and~\ref{tab:BERT_cmp}. Table~\ref{tab:LM_cmp} provides a comprehensive comparison of full fine-tuning versus PEFT approaches on two state-of-the-art decoder-only language models: Qwen2.5-3B and LLaMA3.2-3B. Table~\ref{tab:BERT_cmp} extends this analysis to the encoder-only architecture using DeBERTa-v3-base. From the results, we could find:
\begin{itemize}[topsep=4pt, noitemsep,leftmargin=*]
    \item First, the proposed TASO method demonstrates remarkable parameter efficiency in decoder architectures. TASO achieves the highest average accuracy with only 2.06M and 1.67M trainable parameters on Qwen2.5-3B and LLaMA3.2-3B, respectively---over 30 $\times$ fewer than LoRA (r=32) and DoRA (r=32). On Qwen2.5-3B, TASO outperforms all baselines on GSM8K and BoolQ, while maintaining strong results on WiC and ARC. On LLaMA3.2-3B, TASO achieves top performance on BoolQ and WiC, highlighting its parameter efficiency in decoder-only settings.
    \item Second, In encoder architectures, TASO's efficiency advantage becomes even more pronounced. Requiring only 0.18M parameters (15$\times$ fewer than LoRA with r=16), TASO establishes new state-of-the-art results on the GLUE benchmark. The performance improvements are particularly notable in three critical NLP tasks: textual entailment (RTE: 88.01 vs. LoRA's 87.54), paraphrase detection (MRPC: 90.03 vs. 89.41), and semantic textual similarity (STS-B: 92.30 vs. 91.82). This consistent outperformance across diverse tasks underscores TASO's enhanced generalization capabilities.
\end{itemize}

In summary, these findings confirm that our task-specific sparsification strategy significantly reduces redundancy in LoRA modules while preserving or enhancing performance across a broad range of tasks and model architectures.

\begin{figure*}[t]
  \centering
  \includegraphics[width=0.34\textwidth]{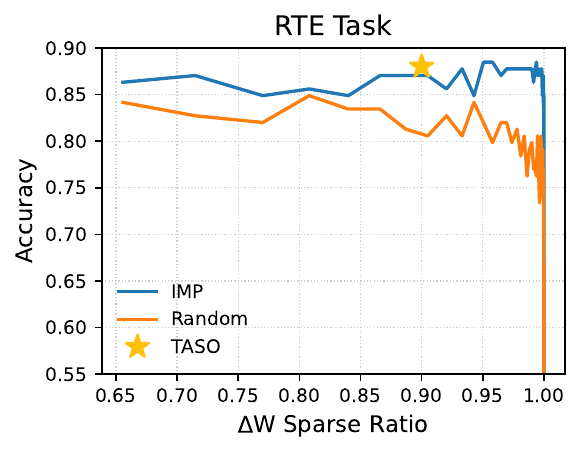}
  \includegraphics[width=0.28\textwidth]{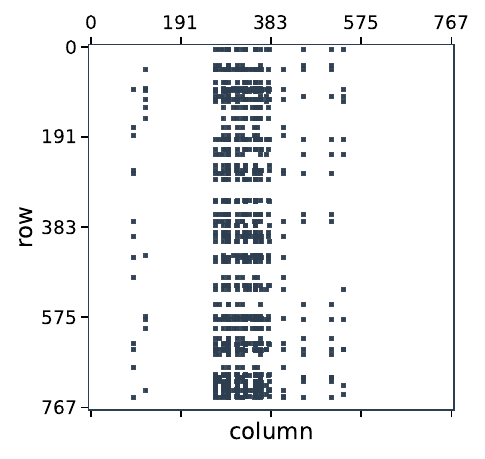}
  \includegraphics[width=0.34\textwidth]{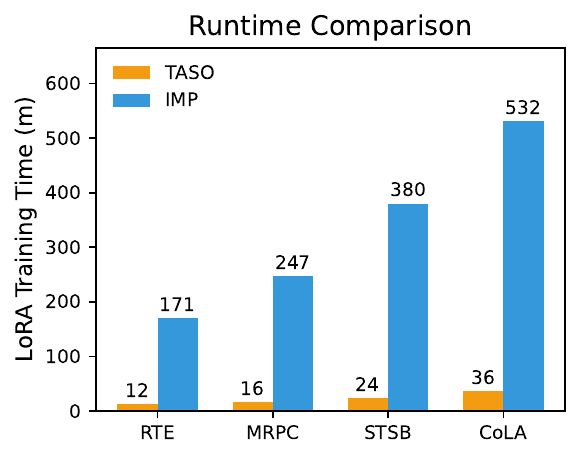}
  \caption{
    \textbf{Left:} Accuracy vs. sparsity curve with TASO highlighted.
    \textbf{Middle:} Visualization of key mask sparsity.
    \textbf{Right:} LoRA training runtime for TASO vs. IMP on four tasks.
  }
  \label{fig:taso_visualization}
\end{figure*}
\subsection{Connections with Lottery Ticket in LoRA}
The \textit{lottery ticket hypothesis} ~\cite{frankle2018lottery} posits that dense neural networks contain sparse, trainable subnetworks capable of matching the performance of the original network when trained in isolation. This phenomenon is typically identified through the Iterative Magnitude Pruning (IMP) algorithm, which progressively removes low-magnitude weights and retrains the remaining subnetwork. Our application of IMP to LoRA modules (r=8) in DeBERTa-v3-base (RTE task) reveals that 99\% sparsity can be achieved while maintaining performance (left figure in Figure~\ref{fig:taso_visualization})- a significant improvement over existing sparse LoRA studies~\cite{ding2023sparselowrankadaptationpretrained,wang2024roselorarowcolumnwisesparse} that typically achieve only 30\% sparsity. This remarkable sparsification potential directly motivates TASO's core design: if such high sparsity is achievable in standard LoRA, then an optimally pruned r=1 LoRA could potentially match the performance while being dramatically more parameter-efficient.

Notably, IMP's surviving weights exhibit column-wise clustering patterns (middle figure in Figure~\ref{fig:taso_visualization}) that align with TASO's task-specific core regions (Figure~\ref{fig:mask_visualization}), suggesting both methods converge to similar sparse configurations through distinct mechanisms.

However, IMP method requires computationally expensive iterative pruning and retraining cycles. In contrast, TASO achieves comparable sparsity levels in two iterative training steps through gradient-informed core region identification. 
The right panel of Figure 3 compares the computational time between TASO and IMP algorithms. Notably, when achieving comparable sparsity levels to TASO, IMP demonstrates approximately an order of magnitude higher computational cost (requiring 171 minutes versus 12 minutes for the RTE task). Furthermore, this performance gap exhibits a pronounced expansion with increasing task complexity. Therefore, it is noteworthy that TASO maintains 90\% sparsity with remarkably low computational overhead.

\begin{table}[t]
\centering
\resizebox{0.95\columnwidth}{!}{
\begin{tabular}{lccc}
\toprule
\textbf{Dataset} & \textbf{TASO} & \textbf{w/o LR scale} & \textbf{w/o core region} \\
\midrule
\multicolumn{4}{c}{(a) LLaMA3.2 3B model} \\
GSM8K      & \textbf{39.81}          & 29.00$_{\color{mygreen}{-10.81}}$ & 37.63$_{\color{mygreen}{-2.18}}$ \\
BoolQ      & \textbf{88.53}          & 80.50$_{\color{mygreen}{-8.03}}$ & 82.14$_{\color{mygreen}{-6.39}}$ \\
WiC        & \textbf{74.50}          & 57.05$_{\color{mygreen}{-17.45}}$ & 61.37$_{\color{mygreen}{-13.13}}$ \\
\hline
\multicolumn{4}{c}{(b) Qwen2.5 3B model} \\
GSM8K      & \textbf{70.04}          & 68.53$_{\color{mygreen}{-1.51}}$ & 67.58$_{\color{mygreen}{-2.46}}$ \\
BoolQ      & \textbf{88.64}          & 86.39$_{\color{mygreen}{-2.25}}$ & 87.19$_{\color{mygreen}{-1.45}}$ \\
WiC        & \textbf{73.33}          & 72.94$_{\color{mygreen}{-0.39}}$ & 72.35$_{\color{mygreen}{-0.98}}$ \\
\hline
\multicolumn{4}{c}{(c) DeBERTa-v3-base model} \\
RTE        & \textbf{88.01}          & 84.17$_{\color{mygreen}{-3.84}}$ & 81.77$_{\color{mygreen}{-6.24}}$ \\
CoLA       & \textbf{69.86}          & 68.33$_{\color{mygreen}{-1.53}}$ & 67.88$_{\color{mygreen}{-1.98}}$ \\
MRPC       & \textbf{90.03}          & 89.71$_{\color{mygreen}{-0.32}}$ & 89.56$_{\color{mygreen}{-0.47}}$ \\
STS-B      & \textbf{92.30}          & 91.93$_{\color{mygreen}{-0.37}}$ & 91.53$_{\color{mygreen}{-0.77}}$ \\
\bottomrule
\end{tabular}
}
\caption{Ablation results across different models and tasks. Both learning rate scaling and core region-based pruning contribute significantly to performance.}
\label{tab:ablation_all}
\end{table}
\subsection{Discussions}
To further inspect the proposed method, we investigate the following research questions.
\paragraph{Does scaled learning rate help sparse training?} To evaluate the necessity of sparsity-aware learning rate scaling, we compare the full TASO framework against a variant with fixed baseline learning rates (\textit{w/o LR scale}) across three model architectures. As shown in Table~\ref{tab:ablation_all}:
\begin{itemize}[topsep=4pt, noitemsep,leftmargin=*]
    \item The impact varies substantially by model architecture, with the LLaMA3.2 3B showing the most dramatic sensitivity: removing learning rate scaling causes performance drops of 10.81, 8.03, and 17.45 points on GSM8K, BoolQ, and WiC respectively. 
    \item The technique proves universally beneficial, with measurable gains across all three model families (LLaMA, Qwen, and DeBERTa) and all seven evaluation tasks, though the magnitude of improvement is task-dependent.
\end{itemize}
These findings suggest that adaptive learning rate adjustment is crucial for effective sparse training, as it helps balance the information flow through the remaining active connections.

\paragraph{Does the identified core region contain critical information?}
We analyze the importance of structured pruning through task-specific core regions by comparing TASO against a variant with randomly selected rows and columns results  (\textit{w/o core region}):
\begin{itemize}[topsep=4pt, noitemsep,leftmargin=*]
    \item Pruning non-core parameters retains >90\% of original performance in most cases (e.g., LLaMA3.2 on GSM8K: 39.81\%→37.63\%), indicating core regions encode compact but sufficient task representations.
    \item The effectiveness holds for both generative (Qwen2.5) and discriminative (DeBERTa) models, suggesting these regions capture architecture-agnostic functional units.
\end{itemize} 
The consistent preservation of functionality confirms that core regions constitute minimal sufficient parameter sets for task execution.

\paragraph{Does Sparse LoRA Improve Cross-Task Compositionality?}
We investigate whether sparsified LoRA modules enhance compositional generalization compared to dense counterparts. The experiment combines:
(1) A base dense LoRA module (fixed, row tasks in Table~\ref{tab:cross_task_comp}), and
(2) Either a standard dense LoRA or our pruned LoRA (column tasks),
evaluated through task-pair accuracy averaging.
Key observations reveal:
\begin{itemize}[topsep=4pt, noitemsep,leftmargin=*]
    \item Pruned modules achieve higher composition accuracy in most task pairs (e.g., CoLA+RTE: 50.40\% vs 41.01\%), demonstrating better cross-task compatibility.
    \item Largest improvements occur when combining semantically distant tasks (MRPC+CoLA: +18.6\%), suggesting sparsity helps isolate task-specific features.
\end{itemize}
These results confirm that sparsifiation not only reduces trainable parameters but also enhances LoRA's compositional properties, providing a principled solution for multi-task adaptation.

Beyond these empirical findings, TASO also offers a broader design perspective for parameter-efficient fine-tuning. Whereas most existing approaches adopt a uniform update strategy across all modules, TASO explicitly leverages gradient-informed importance distributions to identify task-relevant core regions prior to training, and restricts updates to these structures using low-rank LoRA. This task-sensitive and structure-aware strategy not only improves parameter efficiency but also avoids redundant updates to less important components.

Furthermore, our analysis suggests a new interpretation of the role of rank in LoRA: even with extremely low ranks (e.g., r = 1), strong performance can be achieved provided that the mask aligns with the core region. This view goes beyond the intrinsic rank hypothesis, offering new insights for theoretical modeling in PEFT. Importantly, TASO is not mutually exclusive with other approaches; rather, it represents a sparsity-guided and structure-aware design principle that can be combined with complementary techniques. We believe this perspective provides a scalable and composable foundation for future research on extreme parameter-efficient tuning.



\begin{table}[t]
  \centering
  \setlength{\tabcolsep}{4pt}
  \resizebox{0.9\columnwidth}{!}{
  \begin{tabular}{@{}lccc cccc@{}}
    \toprule
     \multirow{2}{*}{\textbf{Task}}     & \multicolumn{3}{c}{\textbf{Dense}}
          & \multicolumn{3}{c}{\textbf{Pruned (TASO)}} \\
    \cmidrule(lr){2-4}\cmidrule(lr){5-7}
     &\textbf{ CoLA} & \textbf{RTE} & \textbf{MRPC} &  \textbf{CoLA} & \textbf{RTE} & \textbf{MRPC} \\
    \midrule
    \textbf{CoLA} & --    & 41.01 & 48.49 & --    & \textbf{50.40} & \textbf{52.87} \\
    \textbf{RTE} & 41.01 & --    & 65.02 & \textbf{60.23} & --    & 63.82 \\
    \textbf{MRPC} & 48.49 & 67.04 & --    & \textbf{67.04} & 59.62 & --    \\
    \bottomrule
  \end{tabular}
  }
  \caption{Cross-task composition accuracy (\%).  
Each row denotes a base task whose dense LoRA module is fixed.  
Each column corresponds to a second LoRA module trained on a different task,  
which is either a standard dense module (left) or a pruned module obtained by our method (right).}
  \label{tab:cross_task_comp}
\end{table}


\section{Conclusion} 
By leveraging the positional characteristics of task-specific core regions and the structural properties of LoRA, TASO enables highly efficient fine-tuning with a parameter budget close to LoRA rank \( r = 1 \).  
It not only reduces the number of trainable parameters, but also consistently outperforms standard LoRA  
across a wide range of tasks and model architectures.  
In future work, we aim to explore the generalization ability of TASO beyond language models,  
particularly in the domains of vision and audio.
\section{Limitations}

While TASO demonstrates promising results, our study has several limitations.
This work evaluates TASO solely on standard NLP tasks, and its applicability to other modalities—such as vision or multimodal models—remains an open question.
Further research is needed to assess whether TASO can generalize as effectively as other PEFT techniques across diverse settings.
Moreover, although TASO offers new insights into parameter-efficient fine-tuning, each component of the method still leaves room for further optimization.

\section*{Acknowledgments}
The authors wish to thank all reviewers for their
helpful comments and suggestions. The corresponding author
is Yuanbin Wu.
This research was (partially) supported by National Key R\&D Program of China (2024YFC3308103).

\bibliography{ref}

@inproceedings{kopiczko2024veravectorbasedrandommatrix,
  author       = {Dawid Jan Kopiczko and
                  Tijmen Blankevoort and
                  Yuki M. Asano},
  title        = {VeRA: Vector-based Random Matrix Adaptation},
  booktitle    = {The Twelfth International Conference on Learning Representations,
                  {ICLR} 2024, Vienna, Austria, May 7-11, 2024},
  publisher    = {OpenReview.net},
  year         = {2024},
  url          = {https://openreview.net/forum?id=NjNfLdxr3A},
  bibsource    = {dblp computer science bibliography, https://dblp.org}
}

@article{ding2023sparselowrankadaptationpretrained,
  author       = {Haoyu Wang and
                  Tianci Liu and
                  Tuo Zhao and
                  Jing Gao},
  title        = {RoseLoRA: Row and Column-wise Sparse Low-rank Adaptation of Pre-trained
                  Language Model for Knowledge Editing and Fine-tuning},
  journal      = {CoRR},
  volume       = {abs/2406.10777},
  year         = {2024},
  url          = {https://doi.org/10.48550/arXiv.2406.10777},
  doi          = {10.48550/ARXIV.2406.10777},
  eprinttype    = {arXiv},
  eprint       = {2406.10777},
  bibsource    = {dblp computer science bibliography, https://dblp.org}
}

@article{zhang2023adaloraadaptivebudgetallocation,
  title={Adalora: Adaptive budget allocation for parameter-efficient fine-tuning},
  author={Zhang, Qingru and Chen, Minshuo and Bukharin, Alexander and Karampatziakis, Nikos and He, Pengcheng and Cheng, Yu and Chen, Weizhu and Zhao, Tuo},
  journal={arXiv preprint arXiv:2303.10512},
  year={2023}
}

@inproceedings{yu2024languagemodelssupermario,
  title={Language models are super mario: Absorbing abilities from homologous models as a free lunch},
  author={Yu, Le and Yu, Bowen and Yu, Haiyang and Huang, Fei and Li, Yongbin},
  booktitle={Forty-first International Conference on Machine Learning},
  year={2024}
}

@article{panda2024lotteryticketadaptationmitigating,
  title={Lottery ticket adaptation: Mitigating destructive interference in llms},
  author={Panda, Ashwinee and Isik, Berivan and Qi, Xiangyu and Koyejo, Sanmi and Weissman, Tsachy and Mittal, Prateek},
  journal={arXiv preprint arXiv:2406.16797},
  year={2024}
}

@inproceedings{zhang-etal-2024-unveiling-linguistic,
    title = "Unveiling Linguistic Regions in Large Language Models",
    author = "Zhang, Zhihao  and
      Zhao, Jun  and
      Zhang, Qi  and
      Gui, Tao  and
      Huang, Xuanjing",
    editor = "Ku, Lun-Wei  and
      Martins, Andre  and
      Srikumar, Vivek",
    booktitle = "Proceedings of the 62nd Annual Meeting of the Association for Computational Linguistics (Volume 1: Long Papers)",
    month = aug,
    year = "2024",
    address = "Bangkok, Thailand",
    publisher = "Association for Computational Linguistics",
    url = "https://aclanthology.org/2024.acl-long.338/",
    doi = "10.18653/v1/2024.acl-long.338",
    pages = "6228--6247"
}

@inproceedings{houlsby2019parameterefficienttransferlearningnlp,
  title={Parameter-efficient transfer learning for NLP},
  author={Houlsby, Neil and Giurgiu, Andrei and Jastrzebski, Stanislaw and Morrone, Bruna and De Laroussilhe, Quentin and Gesmundo, Andrea and Attariyan, Mona and Gelly, Sylvain},
  booktitle={International conference on machine learning},
  pages={2790--2799},
  year={2019},
  organization={PMLR}
}

@article{li2021prefixtuningoptimizingcontinuousprompts,
  title={Prefix-tuning: Optimizing continuous prompts for generation},
  author={Li, Xiang Lisa and Liang, Percy},
  journal={arXiv preprint arXiv:2101.00190},
  year={2021}
}

@inproceedings{li-liang-2021-prefix,
    title = "Prefix-Tuning: Optimizing Continuous Prompts for Generation",
    author = "Li, Xiang Lisa  and
      Liang, Percy",
    editor = "Zong, Chengqing  and
      Xia, Fei  and
      Li, Wenjie  and
      Navigli, Roberto",
    booktitle = "Proceedings of the 59th Annual Meeting of the Association for Computational Linguistics and the 11th International Joint Conference on Natural Language Processing (Volume 1: Long Papers)",
    month = aug,
    year = "2021",
    address = "Online",
    publisher = "Association for Computational Linguistics",
    url = "https://aclanthology.org/2021.acl-long.353/",
    doi = "10.18653/v1/2021.acl-long.353",
    pages = "4582--4597"
}

@article{hu2022sparsestructuresearchparameterefficient,
  title={Sparse structure search for parameter-efficient tuning},
  author={Hu, Shengding and Zhang, Zhen and Ding, Ning and Wang, Yadao and Wang, Yasheng and Liu, Zhiyuan and Sun, Maosong},
  journal={arXiv preprint arXiv:2206.07382},
  year={2022}
}

@inproceedings{guo-etal-2021-parameter,
    title = "Parameter-Efficient Transfer Learning with Diff Pruning",
    author = "Guo, Demi  and
      Rush, Alexander  and
      Kim, Yoon",
    editor = "Zong, Chengqing  and
      Xia, Fei  and
      Li, Wenjie  and
      Navigli, Roberto",
    booktitle = "Proceedings of the 59th Annual Meeting of the Association for Computational Linguistics and the 11th International Joint Conference on Natural Language Processing (Volume 1: Long Papers)",
    month = aug,
    year = "2021",
    address = "Online",
    publisher = "Association for Computational Linguistics",
    url = "https://aclanthology.org/2021.acl-long.378/",
    doi = "10.18653/v1/2021.acl-long.378",
    pages = "4884--4896"
}

@article{luo2024moeloracontrastivelearningguided,
  title={Moelora: Contrastive learning guided mixture of experts on parameter-efficient fine-tuning for large language models},
  author={Luo, Tongxu and Lei, Jiahe and Lei, Fangyu and Liu, Weihao and He, Shizhu and Zhao, Jun and Liu, Kang},
  journal={arXiv preprint arXiv:2402.12851},
  year={2024}
}

@article{huang2024lorahubefficientcrosstaskgeneralization,
  title={Lorahub: Efficient cross-task generalization via dynamic lora composition},
  author={Huang, Chengsong and Liu, Qian and Lin, Bill Yuchen and Pang, Tianyu and Du, Chao and Lin, Min},
  journal={arXiv preprint arXiv:2307.13269},
  year={2023}
}

@inproceedings{shah2023ziplorasubjectstyleeffectively,
  title={Ziplora: Any subject in any style by effectively merging loras},
  author={Shah, Viraj and Ruiz, Nataniel and Cole, Forrester and Lu, Erika and Lazebnik, Svetlana and Li, Yuanzhen and Jampani, Varun},
  booktitle={European Conference on Computer Vision},
  pages={422--438},
  year={2024},
  organization={Springer}
}

@article{tian2024hydraloraasymmetricloraarchitecture,
  title={Hydralora: An asymmetric lora architecture for efficient fine-tuning},
  author={Tian, Chunlin and Shi, Zhan and Guo, Zhijiang and Li, Li and Xu, Cheng-Zhong},
  journal={Advances in Neural Information Processing Systems},
  volume={37},
  pages={9565--9584},
  year={2024}
}

@article{valipour2023dyloraparameterefficienttuning,
  title={Dylora: Parameter efficient tuning of pre-trained models using dynamic search-free low-rank adaptation},
  author={Valipour, Mojtaba and Rezagholizadeh, Mehdi and Kobyzev, Ivan and Ghodsi, Ali},
  journal={arXiv preprint arXiv:2210.07558},
  year={2022}
}

@inproceedings{wang2024roselorarowcolumnwisesparse,
  author       = {Haoyu Wang and
                  Tianci Liu and
                  Ruirui Li and
                  Monica Xiao Cheng and
                  Tuo Zhao and
                  Jing Gao},
  editor       = {Yaser Al{-}Onaizan and
                  Mohit Bansal and
                  Yun{-}Nung Chen},
  title        = {RoseLoRA: Row and Column-wise Sparse Low-rank Adaptation of Pre-trained
                  Language Model for Knowledge Editing and Fine-tuning},
  booktitle    = {Proceedings of the 2024 Conference on Empirical Methods in Natural
                  Language Processing, {EMNLP} 2024, Miami, FL, USA, November 12-16,
                  2024},
  pages        = {996--1008},
  publisher    = {Association for Computational Linguistics},
  year         = {2024},
  url          = {https://aclanthology.org/2024.emnlp-main.57},
  bibsource    = {dblp computer science bibliography, https://dblp.org}
}

@inproceedings{DBLP:conf/emnlp/YaoGLZWWZ24,
  author       = {Kai Yao and
                  Penglei Gao and
                  Lichun Li and
                  Yuan Zhao and
                  Xiaofeng Wang and
                  Wei Wang and
                  Jianke Zhu},
  editor       = {Yaser Al{-}Onaizan and
                  Mohit Bansal and
                  Yun{-}Nung Chen},
  title        = {Layer-wise Importance Matters: Less Memory for Better Performance
                  in Parameter-efficient Fine-tuning of Large Language Models},
  booktitle    = {Findings of the Association for Computational Linguistics: {EMNLP}
                  2024, Miami, Florida, USA, November 12-16, 2024},
  pages        = {1977--1992},
  publisher    = {Association for Computational Linguistics},
  year         = {2024},
  url          = {https://aclanthology.org/2024.findings-emnlp.109},
  bibsource    = {dblp computer science bibliography, https://dblp.org}
}

@inproceedings{DBLP:conf/acl/DaiDHSCW22,
  author       = {Damai Dai and
                  Li Dong and
                  Yaru Hao and
                  Zhifang Sui and
                  Baobao Chang and
                  Furu Wei},
  editor       = {Smaranda Muresan and
                  Preslav Nakov and
                  Aline Villavicencio},
  title        = {Knowledge Neurons in Pretrained Transformers},
  booktitle    = {Proceedings of the 60th Annual Meeting of the Association for Computational
                  Linguistics (Volume 1: Long Papers), {ACL} 2022, Dublin, Ireland,
                  May 22-27, 2022},
  pages        = {8493--8502},
  publisher    = {Association for Computational Linguistics},
  year         = {2022},
  url          = {https://doi.org/10.18653/v1/2022.acl-long.581},
  doi          = {10.18653/V1/2022.ACL-LONG.581},
  bibsource    = {dblp computer science bibliography, https://dblp.org}
}

@inproceedings{DBLP:conf/aaai/HuangHJZC25,
  author       = {Weiyu Huang and
                  Yuezhou Hu and
                  Guohao Jian and
                  Jun Zhu and
                  Jianfei Chen},
  editor       = {Toby Walsh and
                  Julie Shah and
                  Zico Kolter},
  title        = {Pruning Large Language Models with Semi-Structural Adaptive Sparse
                  Training},
  booktitle    = {AAAI-25, Sponsored by the Association for the Advancement of Artificial
                  Intelligence, February 25 - March 4, 2025, Philadelphia, PA, {USA}},
  pages        = {24167--24175},
  publisher    = {{AAAI} Press},
  year         = {2025},
  url          = {https://doi.org/10.1609/aaai.v39i23.34592},
  doi          = {10.1609/AAAI.V39I23.34592},
  bibsource    = {dblp computer science bibliography, https://dblp.org}
}

@article{zhao2023survey,
  title={A survey of large language models},
  author={Zhao, Wayne Xin and Zhou, Kun and Li, Junyi and Tang, Tianyi and Wang, Xiaolei and Hou, Yupeng and Min, Yingqian and Zhang, Beichen and Zhang, Junjie and Dong, Zican and others},
  journal={arXiv preprint arXiv:2303.18223},
  volume={1},
  number={2},
  year={2023}
}

@article{minaee2025largelanguagemodelssurvey,
      author       = {Shervin Minaee and
                  Tom{\'{a}}s Mikolov and
                  Narjes Nikzad and
                  Meysam Chenaghlu and
                  Richard Socher and
                  Xavier Amatriain and
                  Jianfeng Gao},
  title        = {Large Language Models: {A} Survey},
  journal      = {CoRR},
  volume       = {abs/2402.06196},
  year         = {2024},
  url          = {https://doi.org/10.48550/arXiv.2402.06196},
  doi          = {10.48550/ARXIV.2402.06196},
  eprinttype    = {arXiv},
  eprint       = {2402.06196},
  bibsource    = {dblp computer science bibliography, https://dblp.org}
}

@article{zhou2024comprehensive,
  title={A comprehensive survey on pretrained foundation models: A history from bert to chatgpt},
  author={Zhou, Ce and Li, Qian and Li, Chen and Yu, Jun and Liu, Yixin and Wang, Guangjing and Zhang, Kai and Ji, Cheng and Yan, Qiben and He, Lifang and others},
  journal={International Journal of Machine Learning and Cybernetics},
  pages={1--65},
  year={2024},
  publisher={Springer}
}

@article{zhang2024comprehensive,
  title={A comprehensive survey of scientific large language models and their applications in scientific discovery},
  author={Zhang, Yu and Chen, Xiusi and Jin, Bowen and Wang, Sheng and Ji, Shuiwang and Wang, Wei and Han, Jiawei},
  journal={arXiv preprint arXiv:2406.10833},
  year={2024}
}

@article{han2024parameter,
  title={Parameter-efficient fine-tuning for large models: A comprehensive survey},
  author={Han, Zeyu and Gao, Chao and Liu, Jinyang and Zhang, Jeff and Zhang, Sai Qian},
  journal={arXiv preprint arXiv:2403.14608},
  year={2024}
}

@article{ding2023parameter,
  title={Parameter-efficient fine-tuning of large-scale pre-trained language models},
  author={Ding, Ning and Qin, Yujia and Yang, Guang and Wei, Fuchao and Yang, Zonghan and Su, Yusheng and Hu, Shengding and Chen, Yulin and Chan, Chi-Min and Chen, Weize and others},
  journal={Nature Machine Intelligence},
  volume={5},
  number={3},
  pages={220--235},
  year={2023},
  publisher={Nature Publishing Group UK London}
}

@article{wang2024parameter,
  title={Parameter-efficient fine-tuning in large models: A survey of methodologies},
  author={Wang, Luping and Chen, Sheng and Jiang, Linnan and Pan, Shu and Cai, Runze and Yang, Sen and Yang, Fei},
  journal={arXiv preprint arXiv:2410.19878},
  year={2024}
}

@inproceedings{zhou-etal-2024-empirical,
    title = "An Empirical Study on Parameter-Efficient Fine-Tuning for {M}ulti{M}odal Large Language Models",
    author = "Zhou, Xiongtao  and
      He, Jie  and
      Ke, Yuhua  and
      Zhu, Guangyao  and
      Gutierrez Basulto, Victor  and
      Pan, Jeff",
    editor = "Ku, Lun-Wei  and
      Martins, Andre  and
      Srikumar, Vivek",
    booktitle = "Findings of the Association for Computational Linguistics: ACL 2024",
    month = aug,
    year = "2024",
    address = "Bangkok, Thailand",
    publisher = "Association for Computational Linguistics",
    url = "https://aclanthology.org/2024.findings-acl.598/",
    doi = "10.18653/v1/2024.findings-acl.598",
    pages = "10057--10084"
}

@article{prottasha2025peft,
  title={PEFT A2Z: Parameter-Efficient Fine-Tuning Survey for Large Language and Vision Models},
  author={Prottasha, Nusrat Jahan and Chowdhury, Upama Roy and Mohanto, Shetu and Nuzhat, Tasfia and Sami, Abdullah As and Ali, Md Shamol and Sobuj, Md Shohanur Islam and Raman, Hafijur and Kowsher, Md and Garibay, Ozlem Ozmen},
  journal={arXiv preprint arXiv:2504.14117},
  year={2025}
}

@article{hu2022lora,
  title={Lora: Low-rank adaptation of large language models.},
  author={Hu, Edward J and Shen, Yelong and Wallis, Phillip and Allen-Zhu, Zeyuan and Li, Yuanzhi and Wang, Shean and Wang, Lu and Chen, Weizhu and others},
  journal={ICLR},
  volume={1},
  number={2},
  pages={3},
  year={2022}
}

@article{qwen2025qwen25technicalreport,
      author       = {An Yang and
                  Baosong Yang and
                  Beichen Zhang and
                  Binyuan Hui and
                  Bo Zheng and
                  Bowen Yu and
                  Chengyuan Li and
                  Dayiheng Liu and
                  Fei Huang and
                  Haoran Wei and
                  Huan Lin and
                  Jian Yang and
                  Jianhong Tu and
                  Jianwei Zhang and
                  Jianxin Yang and
                  Jiaxi Yang and
                  Jingren Zhou and
                  Junyang Lin and
                  Kai Dang and
                  Keming Lu and
                  Keqin Bao and
                  Kexin Yang and
                  Le Yu and
                  Mei Li and
                  Mingfeng Xue and
                  Pei Zhang and
                  Qin Zhu and
                  Rui Men and
                  Runji Lin and
                  Tianhao Li and
                  Tingyu Xia and
                  Xingzhang Ren and
                  Xuancheng Ren and
                  Yang Fan and
                  Yang Su and
                  Yichang Zhang and
                  Yu Wan and
                  Yuqiong Liu and
                  Zeyu Cui and
                  Zhenru Zhang and
                  Zihan Qiu},
  title        = {Qwen2.5 Technical Report},
  journal      = {CoRR},
  volume       = {abs/2412.15115},
  year         = {2024},
  url          = {https://doi.org/10.48550/arXiv.2412.15115},
  doi          = {10.48550/ARXIV.2412.15115},
  eprinttype    = {arXiv},
  eprint       = {2412.15115},
  bibsource    = {dblp computer science bibliography, https://dblp.org}
}

@article{cobbe2021training,
  title={Training verifiers to solve math word problems},
  author={Cobbe, Karl and Kosaraju, Vineet and Bavarian, Mohammad and Chen, Mark and Jun, Heewoo and Kaiser, Lukasz and Plappert, Matthias and Tworek, Jerry and Hilton, Jacob and Nakano, Reiichiro and others},
  journal={arXiv preprint arXiv:2110.14168},
  year={2021}
}

@article{clark2019boolq,
  title={Boolq: Exploring the surprising difficulty of natural yes/no questions},
  author={Clark, Christopher and Lee, Kenton and Chang, Ming-Wei and Kwiatkowski, Tom and Collins, Michael and Toutanova, Kristina},
  journal={arXiv preprint arXiv:1905.10044},
  year={2019}
}

@article{pilehvar2018wic,
  title={WiC: the word-in-context dataset for evaluating context-sensitive meaning representations},
  author={Pilehvar, Mohammad Taher and Camacho-Collados, Jose},
  journal={arXiv preprint arXiv:1808.09121},
  year={2018}
}

@article{clark2018think,
  title={Think you have solved question answering? try arc, the ai2 reasoning challenge},
  author={Clark, Peter and Cowhey, Isaac and Etzioni, Oren and Khot, Tushar and Sabharwal, Ashish and Schoenick, Carissa and Tafjord, Oyvind},
  journal={arXiv preprint arXiv:1803.05457},
  year={2018}
}

@article{wang2018glue,
  title={GLUE: A multi-task benchmark and analysis platform for natural language understanding},
  author={Wang, Alex and Singh, Amanpreet and Michael, Julian and Hill, Felix and Levy, Omer and Bowman, Samuel R},
  journal={arXiv preprint arXiv:1804.07461},
  year={2018}
}

@article{warstadt2019neural,
  title={Neural Network Acceptability Judgments},
  author={Warstadt, Alex and Singh, Amanpreet and Bowman, Samuel R},
  journal={Transactions of the Association for Computational Linguistics},
  volume={7},
  pages={625--641},
  year={2019}
}

@inproceedings{socher2013recursive,
  title={Recursive Deep Models for Semantic Compositionality Over a Sentiment Treebank},
  author={Socher, Richard and Perelygin, Alex and Wu, Jean and Chuang, Jason and Manning, Christopher D and Ng, Andrew and Potts, Christopher},
  booktitle={Proceedings of the 2013 Conference on Empirical Methods in Natural Language Processing},
  pages={1631--1642},
  year={2013}
}

@inproceedings{dolan2005automatically,
  title={Automatically Constructing a Corpus of Sentential Paraphrases},
  author={Dolan, William B and Brockett, Chris},
  booktitle={Proceedings of the Third International Workshop on Paraphrasing},
  year={2005}
}

@misc{quora2017question,
  title={Quora Question Pairs},
  author={{Quora}},
  year={2017},
  note={\url{https://www.quora.com/q/quoradata/First-Quora-Dataset-Release-Question-Pairs}}
}

@inproceedings{cer2017semeval,
  title={SemEval-2017 Task 1: Semantic Textual Similarity—Multilingual and Cross-lingual Focused Evaluation},
  author={Cer, Daniel and Diab, Mona and Agirre, Eneko and Lopez-Gazpio, Inigo and Specia, Lucia},
  booktitle={Proceedings of the 11th International Workshop on Semantic Evaluation (SemEval-2017)},
  pages={1--14},
  year={2017}
}

@inproceedings{williams2018broad,
  title={A Broad-Coverage Challenge Corpus for Sentence Understanding through Inference},
  author={Williams, Adina and Nangia, Nikita and Bowman, Samuel R},
  booktitle={Proceedings of the 2018 Conference of the North American Chapter of the Association for Computational Linguistics: Human Language Technologies},
  pages={1112--1122},
  year={2018}
}

@inproceedings{rajpurkar2016squad,
  author       = {Pranav Rajpurkar and
                  Jian Zhang and
                  Konstantin Lopyrev and
                  Percy Liang},
  editor       = {Jian Su and
                  Xavier Carreras and
                  Kevin Duh},
  title        = {SQuAD: 100, 000+ Questions for Machine Comprehension of Text},
  booktitle    = {Proceedings of the 2016 Conference on Empirical Methods in Natural
                  Language Processing, {EMNLP} 2016, Austin, Texas, USA, November 1-4,
                  2016},
  pages        = {2383--2392},
  publisher    = {The Association for Computational Linguistics},
  year         = {2016},
  url          = {https://doi.org/10.18653/v1/d16-1264},
  doi          = {10.18653/V1/D16-1264},
  bibsource    = {dblp computer science bibliography, https://dblp.org}
}

@inproceedings{dagan2005pascal,
  title={The PASCAL Recognising Textual Entailment Challenge},
  author={Dagan, Ido and Glickman, Oren and Magnini, Bernardo},
  booktitle={Machine Learning Challenges Workshop},
  pages={177--190},
  year={2005}
}

@inproceedings{he2021debertav3,
  author       = {Pengcheng He and
                  Jianfeng Gao and
                  Weizhu Chen},
  title        = {DeBERTaV3: Improving DeBERTa using ELECTRA-Style Pre-Training with
                  Gradient-Disentangled Embedding Sharing},
  booktitle    = {The Eleventh International Conference on Learning Representations,
                  {ICLR} 2023, Kigali, Rwanda, May 1-5, 2023},
  publisher    = {OpenReview.net},
  year         = {2023},
  url          = {https://openreview.net/forum?id=sE7-XhLxHA},
  bibsource    = {dblp computer science bibliography, https://dblp.org}
}

@article{grattafiori2024llama3herdmodels,
      author       = {Abhimanyu Dubey and
                  Abhinav Jauhri and
                  Abhinav Pandey and
                  Abhishek Kadian and
                  Ahmad Al{-}Dahle and
                  Aiesha Letman and
                  Akhil Mathur and
                  Alan Schelten and
                  Amy Yang and
                  Angela Fan and
                  Anirudh Goyal and
                  Anthony Hartshorn and
                  Aobo Yang and
                  Archi Mitra and
                  Archie Sravankumar and
                  Artem Korenev and
                  Arthur Hinsvark and
                  Arun Rao and
                  Aston Zhang and
                  Aur{\'{e}}lien Rodriguez and
                  Austen Gregerson and
                  Ava Spataru and
                  Baptiste Rozi{\`{e}}re and
                  Bethany Biron and
                  Binh Tang and
                  Bobbie Chern and
                  Charlotte Caucheteux and
                  Chaya Nayak and
                  Chloe Bi and
                  Chris Marra and
                  Chris McConnell and
                  Christian Keller and
                  Christophe Touret and
                  Chunyang Wu and
                  Corinne Wong and
                  Cristian Canton Ferrer and
                  Cyrus Nikolaidis and
                  Damien Allonsius and
                  Daniel Song and
                  Danielle Pintz and
                  Danny Livshits and
                  David Esiobu and
                  Dhruv Choudhary and
                  Dhruv Mahajan and
                  Diego Garcia{-}Olano and
                  Diego Perino and
                  Dieuwke Hupkes and
                  Egor Lakomkin and
                  Ehab AlBadawy and
                  Elina Lobanova and
                  Emily Dinan and
                  Eric Michael Smith and
                  Filip Radenovic and
                  Frank Zhang and
                  Gabriel Synnaeve and
                  Gabrielle Lee and
                  Georgia Lewis Anderson and
                  Graeme Nail and
                  Gr{\'{e}}goire Mialon and
                  Guan Pang and
                  Guillem Cucurell and
                  Hailey Nguyen and
                  Hannah Korevaar and
                  Hu Xu and
                  Hugo Touvron and
                  Iliyan Zarov and
                  Imanol Arrieta Ibarra and
                  Isabel M. Kloumann and
                  Ishan Misra and
                  Ivan Evtimov and
                  Jade Copet and
                  Jaewon Lee and
                  Jan Geffert and
                  Jana Vranes and
                  Jason Park and
                  Jay Mahadeokar and
                  Jeet Shah and
                  Jelmer van der Linde and
                  Jennifer Billock and
                  Jenny Hong and
                  Jenya Lee and
                  Jeremy Fu and
                  Jianfeng Chi and
                  Jianyu Huang and
                  Jiawen Liu and
                  Jie Wang and
                  Jiecao Yu and
                  Joanna Bitton and
                  Joe Spisak and
                  Jongsoo Park and
                  Joseph Rocca and
                  Joshua Johnstun and
                  Joshua Saxe and
                  Junteng Jia and
                  Kalyan Vasuden Alwala and
                  Kartikeya Upasani and
                  Kate Plawiak and
                  Ke Li and
                  Kenneth Heafield and
                  Kevin Stone and
                  et al.},
  title        = {The Llama 3 Herd of Models},
  journal      = {CoRR},
  volume       = {abs/2407.21783},
  year         = {2024},
  url          = {https://doi.org/10.48550/arXiv.2407.21783},
  doi          = {10.48550/ARXIV.2407.21783},
  eprinttype    = {arXiv},
  eprint       = {2407.21783},
  bibsource    = {dblp computer science bibliography, https://dblp.org}
}

@inproceedings{ben-zaken-etal-2022-bitfit,
    title = "{B}it{F}it: Simple Parameter-efficient Fine-tuning for Transformer-based Masked Language-models",
    author = "Ben Zaken, Elad  and
      Goldberg, Yoav  and
      Ravfogel, Shauli",
    editor = "Muresan, Smaranda  and
      Nakov, Preslav  and
      Villavicencio, Aline",
    booktitle = "Proceedings of the 60th Annual Meeting of the Association for Computational Linguistics (Volume 2: Short Papers)",
    month = may,
    year = "2022",
    address = "Dublin, Ireland",
    publisher = "Association for Computational Linguistics",
    url = "https://aclanthology.org/2022.acl-short.1/",
    doi = "10.18653/v1/2022.acl-short.1",
    pages = "1--9"
}

@article{liu2022few,
  title={Few-shot parameter-efficient fine-tuning is better and cheaper than in-context learning},
  author={Liu, Haokun and Tam, Derek and Muqeeth, Mohammed and Mohta, Jay and Huang, Tenghao and Bansal, Mohit and Raffel, Colin A},
  journal={Advances in Neural Information Processing Systems},
  volume={35},
  pages={1950--1965},
  year={2022}
}

@article{frankle2018lottery,
  title={The lottery ticket hypothesis: Finding sparse, trainable neural networks},
  author={Frankle, Jonathan and Carbin, Michael},
  journal={arXiv preprint arXiv:1803.03635},
  year={2018}
}

@inproceedings{DBLP:conf/iclr/KopiczkoBA24,
  author       = {Dawid Jan Kopiczko and
                  Tijmen Blankevoort and
                  Yuki M. Asano},
  title        = {VeRA: Vector-based Random Matrix Adaptation},
  booktitle    = {The Twelfth International Conference on Learning Representations,
                  {ICLR} 2024, Vienna, Austria, May 7-11, 2024},
  publisher    = {OpenReview.net},
  year         = {2024},
  url          = {https://openreview.net/forum?id=NjNfLdxr3A},
  bibsource    = {dblp computer science bibliography, https://dblp.org}
}

@article{DBLP:journals/corr/abs-2405-17604,
  author       = {Klaudia Balazy and
                  Mohammadreza Banaei and
                  Karl Aberer and
                  Jacek Tabor},
  title        = {LoRA-XS: Low-Rank Adaptation with Extremely Small Number of Parameters},
  journal      = {CoRR},
  volume       = {abs/2405.17604},
  year         = {2024},
  url          = {https://doi.org/10.48550/arXiv.2405.17604},
  doi          = {10.48550/ARXIV.2405.17604},
  eprinttype    = {arXiv},
  eprint       = {2405.17604},
  bibsource    = {dblp computer science bibliography, https://dblp.org}
}

\appendix

\section{Comparison with different importance scores}
\label{sec:importance_score}

\begin{table}[t]
\centering

\resizebox{\columnwidth}{!}{         
\begin{tabular}{cllll}
\toprule
\textbf{Method} & \textbf{Model} & \textbf{GSM8K} & \textbf{BoolQ} &\textbf{ WiC} \\
\midrule
sensitivity      & \multirow{2}{*}{\centering Qwen2.5}  & \textbf{70.04} & \textbf{88.64} & 73.33 \\
gradient  &                                          & 67.67$_{\color{mygreen}{-2.37}}$ & 88.53$_{\color{mygreen}{-0.11}}$ & 73.33$_{\color{mygreen}{-0.00}}$ \\
\midrule
sensitivity      & \multirow{2}{*}{\centering LLaMA3.2} & \textbf{39.81} & \textbf{88.53} & \textbf{74.50} \\
gradient &                                          & 39.43$_{\color{mygreen}{-0.38}}$ & 87.00$_{\color{mygreen}{-1.53}}$ & 69.01$_{\color{mygreen}{-5.49}}$ \\
\bottomrule
\end{tabular}
}
\caption{Comparison of importance metrics for TASO.  
sensitivity uses $|\theta_j \cdot g_j|$, while gradient uses $|g_j|$ only.}
\label{tab:grad_sum_compare}
\end{table}

We evaluate how different parameter importance metrics affect model performance by comparing TASO's sensitivity scoring $|\theta_j \cdot g_j|$ against gradient-only scoring $|g_j|$. 
Experiments are conducted on Qwen2.5 3B and LLaMA3.2 3B across three reasoning tasks (GSM8K, BoolQ, WiC). 
Results (in Table~\ref{tab:grad_sum_compare}) show that the joint metric performs matches or outperforms than gradient-only scoring, with particularly strong gains on LLaMA3.2's WiC task (+5.49\%).
These results demonstrate that incorporating parameter magnitude alongside gradients provides more reliable importance estimates.

\begin{figure}[t]
  \centering
  \includegraphics[width=0.48\textwidth]{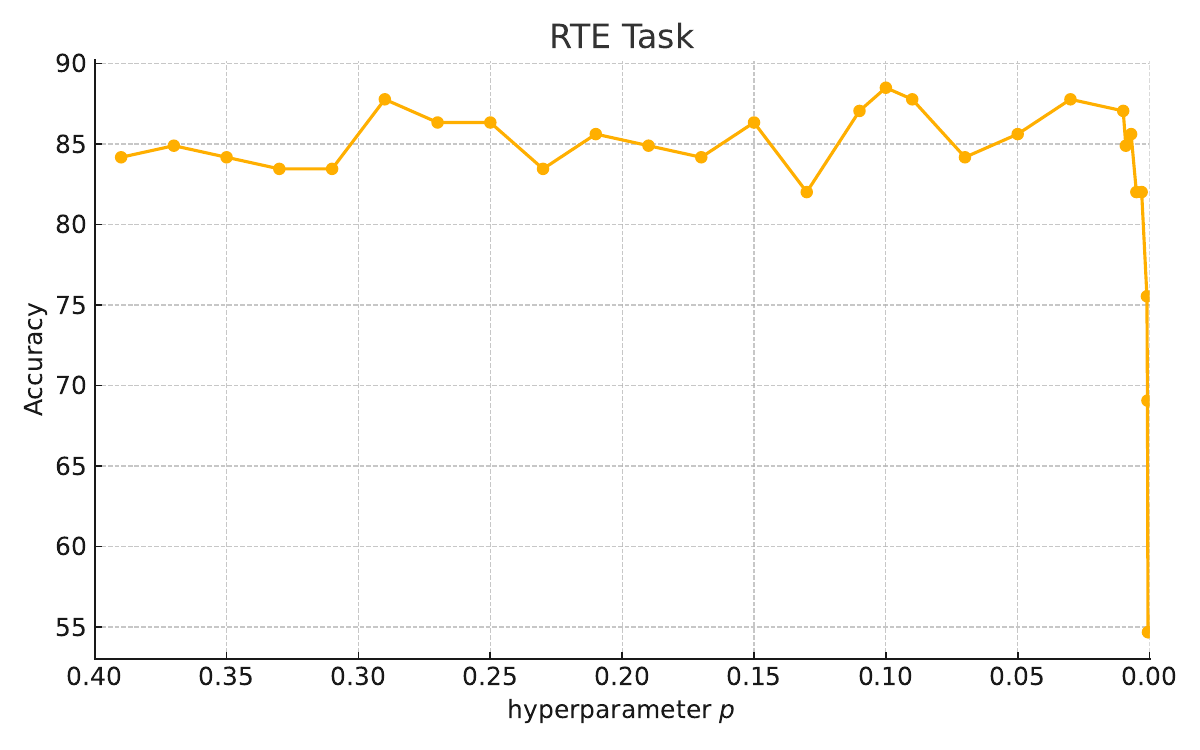}
  \caption{
    Accuracy on the RTE task as a function of the pruning hyperparameter~$p$, which indicates the fraction of  non-zero values after pruned.
    The $x$-axis is shown from $0.40$ (left) to $0$ (right) to highlight
    performance under increasing sparsity.
  }
  \label{fig:rte_accuracy_vs_hp}
\end{figure}

\section{Impact of sparsity ratio on Performance}
\label{sec:impact_p}
We conduct a systematic analysis of the sparsity hyperparameter $p$ (fraction of retained parameters) using the RTE task, with results visualized in Figure~\ref{fig:rte_accuracy_vs_hp}. Our experiments reveal three distinct operational regimes:
\begin{itemize}[topsep=4pt, noitemsep,leftmargin=*]
    \item Remarkably maintains performance despite retaining only 2-5\% parameters, demonstrating TASO's ability to preserve critical information under extreme sparsity conditions. This suggests our importance scoring effectively identifies and protects task-essential parameters.
    \item Across a wide sparsity range (from 5\% to 40\% retained parameters), the method sustains over 90\% of peak accuracy, confirming that most parameters in standard LoRA modules are indeed redundant for task adaptation.
    \item The sweet spot ($p\!=\!0.1$) achieves peak performance (approximately 88\%) while reducing parameters by 90\%, striking the ideal balance between compactness and capability.
\end{itemize}
These findings demonstrate that TASO's structured pruning approach effectively identifies and preserves task-essential parameters, enabling radical parameter reduction without compromising model effectiveness. The consistent performance across sparsity levels suggests our method's suitability for both memory-constrained and high-performance scenarios.

\section{Additional Experimental Details}

\subsection{Learning Rate Settings}
\label{appendix:learning-rates}

For Qwen2.5-3B, the learning rate configurations of different fine-tuning methods are summarized in Table~\ref{tab:learning-rates}.

\begin{table}[h]
\centering
\begin{tabular}{lc}
\hline
\textbf{Method} & \textbf{Learning Rate} \\
\hline
\textbf{TASO / LoRA} & $5\times 10^{-5}$ \\
Full Finetune & $5\times 10^{-6}$ \\
DoRA & $5\times 10^{-5}$ \\
Adapter & $5\times 10^{-5}$ \\
BitFit & $3\times 10^{-3}$ \\
IA$^{3}$ & $3\times 10^{-3}$ \\
LoRA-XS & $1\times 10^{-4}$ \\
VeRA & $1\times 10^{-2}$ \\
\hline
\end{tabular}
\caption{Learning rate configurations for different fine-tuning methods on Qwen2.5-3B.}
\label{tab:learning-rates}
\end{table}

\noindent\textbf{Note.} In practice, we initialize the learning rate with the same setting as LoRA, but during training the code dynamically adjusts it according to the method described in Section~4.3.

\end{document}